\pdfoutput=1 

\documentclass[letterpaper, 10 pt, conference]{ieeeconf}  

\IEEEoverridecommandlockouts                              

\usepackage[natbib=true]{biblatex}

\usepackage{pgfplots}
\usepgfplotslibrary{groupplots,dateplot}
\usetikzlibrary{patterns,shapes.arrows}
\pgfplotsset{compat=newest}
\usepgfplotslibrary{groupplots}
\usepgfplotslibrary{fillbetween}

\usepackage{amsfonts}

\usepackage{tikz}
\usetikzlibrary{shapes, arrows.meta, positioning, bending, graphs, graphs.standard}
\usetikzlibrary{arrows,decorations.pathmorphing,positioning,fit,trees,shapes,shadows,automata,calc} 
\usetikzlibrary{patterns,arrows,arrows.meta,calc,shapes,shadows,decorations.pathmorphing,decorations.pathreplacing,automata,shapes.multipart,positioning,shapes.geometric,fit,circuits,trees,shapes.gates.logic.US,fit, matrix,arrows.meta, quotes}
\usetikzlibrary{backgrounds,scopes,spy}
\usepackage{circuitikz}

\usepackage{subfiles}
\usepackage{xcolor}
\usepackage[hidelinks]{hyperref}

\usepackage{neuralnetwork}
\usepackage{listofitems} 

\usepackage{wrapfig}

\usepackage{adjustbox}
\usepackage{multirow}
\usepackage{pbox}
\usepackage{amssymb}

\pgfplotsset{compat=newest}
\def\addlegendimage{\csname pgfplots@addlegendimage\endcsname}

\newcommand{\E}{
    \begin{bmatrix}
        A_C & 0 & 0 & 0 \\
        0 & I & 0 & 0 \\
        0 & 0 & 0 & 0 \\
        0 & 0 & 0 & 0
    \end{bmatrix}
}

\newcommand{\x}{
    \begin{bmatrix}
        q_C \\
        \phi_L \\
        e \\
        j_V
    \end{bmatrix}
}

\newcommand{\J}{
    \begin{bmatrix}
        0 & -A_L & 0 & -A_V \\
        A_L^T & 0 & 0 & 0 \\
        0 & 0 & 0 & 0 \\
        A_V^T & 0 & 0 & 0
    \end{bmatrix}
}

\newcommand{\z}{
    \begin{bmatrix}
        e \\
        \nabla H(\phi_L) \\
        q(q_C) \\
        j_V
    \end{bmatrix}
}

\newcommand{\res}{
    \begin{bmatrix}
        A_R g(A_R^T e) \\
        0 \\
        A_C^T e - q(q_C) \\
        0
    \end{bmatrix}
}

\newcommand{\B}{
    \begin{bmatrix}
        -A_I & 0 \\
        0 & 0 \\
        0 & 0 \\
        0 & -I
    \end{bmatrix}
}

\newcommand{\control}{
    \begin{bmatrix}
        i(t) \\
        v(t)
    \end{bmatrix}
}

\newcommand{\xc}{
    \begin{bmatrix}
        q_C \\ \phi_L \\ e \\ j_V \\ \lambda
    \end{bmatrix}
}

\newcommand{\zc}{
    \begin{bmatrix}
        e \\ \nabla H_{\theta}(\phi_L) \\ q_{\theta}(q_C) \\ j_V \\ \lambda
    \end{bmatrix}
}

\newcommand{\Ec}{
    \begin{bmatrix}
        A_C & 0 & 0 & 0 & 0 \\
        0 & I & 0 & 0 & 0 \\
        0 & 0 & 0 & 0 & 0 \\
        0 & 0 & 0 & 0 & 0 \\
        0 & 0 & 0 & 0 & 0
    \end{bmatrix}
}

\newcommand{\Jc}{
    \begin{bmatrix}
        0 & -A_L & 0 & -A_V & -A_{\lambda} \\
        A_L^T & 0 & 0 & 0 & 0 \\
        0 & 0 & 0 & 0 & 0 \\
        A_V^T & 0 & 0 & 0 & 0 \\
        A_{\lambda}^T & 0 & 0 & 0 & 0
    \end{bmatrix}
}

\newcommand{\resc}{
    \begin{bmatrix}
        A_R g_{\theta}(A_R^T e) \\
        0 \\
        A_C^T e - q_{\theta}(q_C) \\
        0 \\
        0
    \end{bmatrix}
}

\newcommand{\Bc}{
    \begin{bmatrix}
        -A_I & 0 \\
        0 & 0 \\
        0 & 0 \\
        0 & -I \\
        0 & 0
    \end{bmatrix}
}

\newcommand{\controlc}{
    \begin{bmatrix}
        i(t) \\
        v(t)
    \end{bmatrix}
}

\newcommand{\NODE}{
    {\mathrm{NODE}}
}

\definecolor{changes}{HTML}{5B5F97}
\definecolor{changes}{HTML}{000000}
\definecolor{submodel1Color}{HTML}{E05F15}
\definecolor{submodel2Color}{HTML}{07742D}
\definecolor{submodel3Color}{HTML}{4F359B}
\definecolor{compositeModelColor}{HTML}{519DFD}
\definecolor{trueModelColor}{HTML}{130303}

\definecolor{lightSubmodel1Color}{HTML}{F19B6A}
\definecolor{lightSubmodel2Color}{HTML}{0cd452}
\definecolor{lightSubmodel3Color}{HTML}{a494c4}
\definecolor{lightCompositeModelColor}{HTML}{7DB5FC}

\definecolor{backgroundColor}{HTML}{b5c1d3}

\title{\LARGE \bf
Neural Port-Hamiltonian Differential Algebraic Equations for \\ Compositional Learning of Electrical Networks
}

\author{%
 Cyrus Neary$^{\dagger, 1}$, Nathan Tsao$^{\dagger, 2}$, Ufuk Topcu$^{2}$
 \thanks{$^{\dagger}$Indicates equal contribution. Authors listed alphabetically.} 
 \thanks{
 $^{1}$The University of British Columbia.
 }%
\thanks{
$^{2}$ The University of Texas at Austin.
}
\thanks{Contact: cyrus.neary@ubc.ca, \{nathan.tsao, utopcu\}@utexas.edu}
\thanks{
This work was supported in part by ONR N00014-21-1-2164, NSF 2409535, and NSF 2214939.
}
}

\begin{document}

\pgfdeclarelayer{background2}
\pgfdeclarelayer{background1}
\pgfdeclarelayer{foreground1}
\pgfdeclarelayer{foreground2}
\pgfsetlayers{background2,background1,main,foreground1,foreground2}

\maketitle

\begin{abstract}%
We develop compositional learning algorithms for coupled dynamical systems, with a particular focus on electrical networks.
While deep learning has proven effective at modeling complex relationships from data, compositional couplings between system components typically introduce algebraic constraints on state variables, posing challenges to many existing data-driven approaches to modeling dynamical systems. 
Towards developing deep learning models for constrained dynamical systems, we introduce \textit{neural port-Hamiltonian differential algebraic equations} (N-PHDAEs), which use neural networks to parameterize unknown terms in both the differential and algebraic components of a port-Hamiltonian DAE. 
To train these models, we propose an algorithm that uses automatic differentiation to perform index reduction, automatically transforming the neural DAE into an equivalent system of \textit{neural ordinary differential equations} (N-ODEs), for which established model inference and backpropagation methods exist.
Experiments simulating the dynamics of nonlinear circuits exemplify the benefits of our approach: the proposed N-PHDAE model achieves an order of magnitude improvement in prediction accuracy and constraint satisfaction when compared to a baseline N-ODE over long prediction time horizons.
We also validate the compositional capabilities of our approach through experiments on a simulated DC microgrid: we train individual N-PHDAE models for separate grid components, before coupling them to accurately predict the behavior of larger-scale networks.
\end{abstract}


\section{Introduction}
\label{sec:intro}
Many physical systems, such as electrical networks, chemical reaction networks, and multi-body mechanical systems, comprise many interacting subsystems.
Such systems not only exhibit complex dynamics, but are often subject to algebraic constraints that enforce compositional relationships between the subsystems (e.g., energy balance, conservation laws, or geometric couplings).

Deep learning methods that use physics-inspired architectures and training losses have shown promise in learning data-efficient models of \textit{unconstrained} dynamical systems \citep{raissi2019physics,chen2018neural,djeumou2021neural,djeumou2023learn}.
However, such methods are currently unable to learn models that respect the aforementioned algebraic constraints.
This limitation renders compositional approaches to modeling challenging, acting as a barrier to the application of deep-learning-based models for the prediction and control of many real-world systems.
For example, the complexity that arises from large numbers of interacting components can make monolithic approaches to learning system models intractable.
Furthermore, in many applications, system-level data may not be available for training by a single algorithm.
Moreover, the inability of existing deep learning methods to enforce critical constraints can result in models that are not interpretable or robust, and prone to failure in scenarios outside the training data distribution.

Towards addressing these limitations, we introduce \textit{Neural Port-Hamiltonian Differential Algebraic Equations} (N-PHDAEs), a physics-informed and compositional deep learning approach to modeling dynamical systems subject to algebraic constraints. 
In this work, we focus on learning models of electrical networks.
{\color{changes} However, the proposed framework and overall approach may also be adapted to learn control-oriented models of dynamical systems that can be formulated as a broad class of PHDAE.}

\begin{figure*}
    \centering
    \includegraphics[]{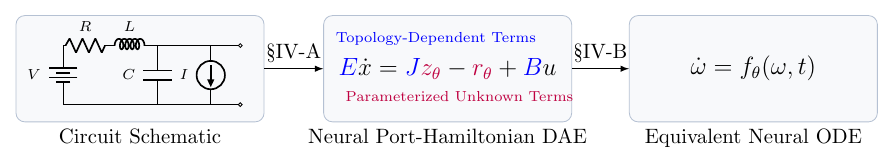}
    \caption{
    The proposed neural port-Hamiltonian differential algebraic equations (N-PHDAE).
    The proposed algorithm uses the topology of the subsystem interconnections to automatically construct a differential algebraic equation (DAE) whose unknown terms are parameterized using neural networks (\S\ref{sec:4.1}).
    It then transforms the resulting neural DAE into an equivalent system of neural ordinary differential equations (\S \ref{sec:4.2}) for inference and training (\S \ref{sec:4.3}).
    }
    \label{fig:PHNDAE}
\end{figure*}

Figure \ref{fig:PHNDAE} illustrates the proposed approach.
The method begins by using a graph describing the interconnection topology of the system's various components to automatically construct the interconnection terms of a Port-Hamiltonian Differential Algebraic Equation (PHDAE) \citep{duindam2009modeling, van2014port, mehrmann2019structure, gunther2020dynamic}.
The remaining terms in the PHDAE (capturing the nonlinear dynamics of the circuit components) are parameterized using neural networks (\S \ref{sec:4.1}).
We then use automatic differentiation to transform the resulting neural DAE into a system of Neural Ordinary Differential Equations (N-ODEs)  (\S \ref{sec:4.2}), which can be more easily trained and evaluated (\S \ref{sec:4.3}).

\begin{figure*}[t]
    \centering
    \vspace{2mm}
    \includegraphics[]{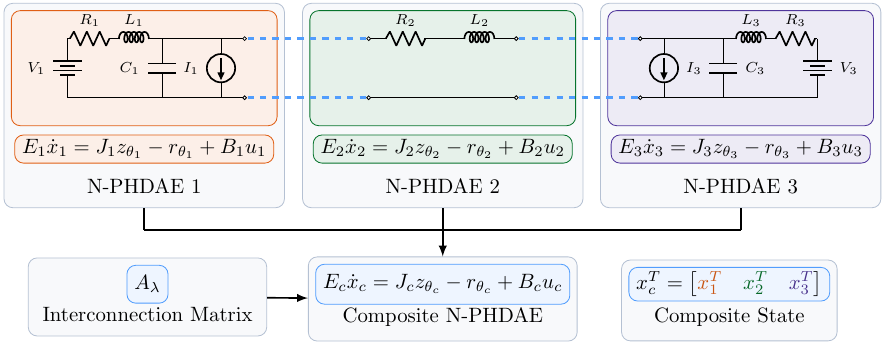}
    \caption{
    The compositionality of the proposed N-PHDAEs (\S \ref{sec:5}).
    By coupling individually trained N-PHDAE models of subsystem dynamics, we obtain a composite N-PHDAE model that generates accurate predictions of more complex system dynamics, such as the DC microgrid example illustrated above and discussed further in \S\ref{sec:6}.
    }
    \label{fig:Comp}
\end{figure*}

To learn \textit{compositional} models of electrical networks, we additionally propose a framework and algorithms to compose N-PHDAEs by coupling their inputs and outputs  (\S \ref{sec:5}), as illustrated in Figure \ref{fig:Comp}.
Separate N-PHDAEs are trained on data generated by individual subsystems.
The trained models are then coupled through an \textit{interconnection matrix} that defines the input-output relationships at the circuit nodes where the couplings occur.
The result is a composite N-PHDAE that models the overall system.

We demonstrate the advantages of the N-PHDAE through case studies on simulated electrical networks.
A N-PHDAE model of the nonlinear FitzHugh-Nagumo circuit demonstrates the approach's data efficiency and accuracy, as well as its ability to learn to satisfy the system's algebraic equations an order of magnitude more accurately than the considered baseline approach---a N-ODE that does not leverage prior physics knowledge or explicitly account for algebraic equations.
Next, we showcase N-PHDAE's compositional modeling capabilities by using it to simulate a DC microgrid.
More specifically, we train ten N-PHDAE models of distributed generation units (DGU) separately, before composing the learned models in previously unseen configurations to accurately simulate larger electrical networks.

\section{Related Work} \label{sec:2}
Methods that leverage physics knowledge in deep learning algorithms have been studied extensively in recent years.
For example, \cite{lu2021physics, raissi2019physics, raissi2018deep, karniadakis2021physics, han2018solving, sirignano2018dgm, long2019pde} use neural networks and deep learning training algorithms to solve partial differential equations.
By contrast, our work focuses on the problem of system identification---learning unknown dynamics from time series data.
More closely related to our work, Neural Ordinary Differential Equations (N-ODEs) are a family of deep learning architectures that use neural networks to parameterize the right-hand side of ODEs.
Originally proposed in the context of generative modeling \cite{chen2018neural}, 
N-ODEs provide a flexible framework for incorporating prior physics knowledge with data-driven models of dynamical systems, yielding models with improved data efficiency and generalization capabilities, especially when training data is limited \citep{djeumou2021neural, djeumou2023learn, rackauckas2020universal, kidger2022neural, neary2024engineering, zhong2021benchmarking,greydanus2019hamiltonian}.
However, N-ODEs struggle to model constrained dynamical systems, often relying on penalty-based methods to enforce known algebraic equations. 

In particular, without significant modification, N-ODEs are unable to model systems that include algebraic constraints on the state variables.
Such DAEs---systems of both differential and algebraic equations---arise frequently in the modeling of physical systems, especially when accounting for couplings between distinct subsystems.
Recently, several approaches to learning DAEs from data have been proposed \citep{xiao2022feasibility,moya2023dae,huang2024minn,koch2024neural}.
These methods rely on learned predictors for the algebraic states or latent variables, which are then used to integrate the DAE.
However, such learned predictors are often trained to fit the available data directly, without leveraging prior physics-based inductive biases that could enhance data efficiency and generalization.
By contrast, we directly incorporate algebraic constraints into the neural network architecture and inference procedure, enabling the training of constraint-respecting and data-efficient models of compositional dynamical systems.

Closely related to our work, \cite{desai2021port,neary2023compositional,tan2024physics,duong2024Port,beckers2023data} also develop algorithms that leverage the mathematical structures defining port-Hamiltonian (PH) systems to learn dynamics models that enjoy key PH properties, e.g., passivity.
Meanwhile, \cite{xu2021neural, furieri2022distributed, plaza2022total} use neural networks to parameterize controllers for systems with known port-Hamiltonian dynamics, and
\cite{neary2023compositional} focuses on leveraging the properties of PH systems to build compositional learning algorithms.
We extend this work by developing port-Hamiltonian neural networks that explicitly account for algebraic equations, {\color{changes} enabling more intuitive compositional coupling of subsystem models.} 

\section{Background}
\label{sec:background}
\paragraph{Port-Hamiltonian Differential Algebraic Equations}
The port-Hamiltonian (PH) framework enables compositional approaches to modeling complex, interconnected systems in a structured and modular way.
Conceptually, the dynamics of individual PH systems are governed by the system's Hamiltonian function $H$, energy dissipation terms, and control inputs.
Separate PH systems can be coupled via energy exchanges through so-called port variables to obtain new PH systems that represent the dynamics of larger composite systems.
We refer to \cite{van2013port,van2014port} for further details.

In this work, we consider port-Hamiltonian differential algebraic equations (PHDAEs)---a broad class of PH systems that include both differential and algebraic equations, and can be written as
\begin{align}
    \frac{d}{dt} E x(t) =  J z(x(t)) - r(z(x(t))) + B u(t).
    \label{eq:PHDescriptor}
\end{align} 
Here $x \in \mathbb{R}^{n}$ is the system's state, $z : \mathbb{R}^{n} \to \mathbb{R}^{n}$ is the effort (a vector-valued function that includes gradients of the system's Hamiltonian function $H(x)$), $E \in \mathbb{R}^{n \times n}$ is the flow matrix, $J \in \mathbb{R}^{n \times n}$ is the skew-symmetric interconnection matrix, $r : \mathbb{R}^{n} \to \mathbb{R}^{n}$ is the dissipation term, $B \in \mathbb{R}^{n \times m}$ is the port matrix, and $u \in \mathbb{R}^m$ is the control input \citep{mehrmann2019structure}.

\paragraph{Port-Hamiltonian Differential Algebraic Equations for Electrical Networks}
Electrical networks may be modeled in PHDAE form as follows \citep{gunther2020dynamic}.
\begin{align}
    \begin{split}
        \frac{d}{dt} \E \x = \\ \J \z \\ - \res + \B \control
    \end{split}
    \label{eq:PHDAE}
\end{align}
We note that \eqref{eq:PHDAE} follows the form of \eqref{eq:PHDescriptor}, where the state $x = \begin{bmatrix} q_C^T & \phi_L^T & e^T & j_V^T \end{bmatrix} \in \mathbb{R}^n$ is composed of the capacitor charges $q_C \in \mathbb{R}^{n_C}$, inductor magnetic fluxes $\phi_L \in \mathbb{R}^{n_L}$, nodal voltages excluding ground $e \in \mathbb{R}^{n_v}$, and current across voltage sources $j_V \in \mathbb{R}^{n_V}$.
The port-Hamiltonian system matrices $E$, $J$, and $B$ depend on incidence matrices $(A_C,A_R,A_L,A_V,A_I)$, where each $A_s \in \{-1,0,1\}^{n_v \times n_s}$ denotes the incidence matrices associated with the capacitors, resistors, inductors, voltage sources, and current sources, respectively. 
Here, $n_s$ denotes the number of elements of each component type. 
The effort function $z$ comprises the resistor voltage-current relation $g: \mathbb{R}^{n_R} \to \mathbb{R}^{n_R}$, capacitor voltage-charge relation $q: \mathbb{R}^{n_C} \to \mathbb{R}^{n_C}$, and the Hamiltonian function $H: \mathbb{R}^{n_L} \to \mathbb{R}$.
The system input $u(t) = \begin{bmatrix} i(t)^T & v(t)^T \end{bmatrix}^T \in \mathbb{R}^{m}$ is comprised of the time-dependent magnitudes of the current and voltage sources. 

\paragraph{Semi-Explicit, Index-1 Differential Algebraic Equations}
More generally, the training algorithms we propose apply to any \textit{index-1} DAEs that may be expressed in the following \textit{semi-explicit} form:
\begin{align}
    \begin{split}
        \dot{v} = f(v,w,t), \qquad 0 = h(v,w,t).
    \end{split}
    \label{eq:DAE}
\end{align}
That is, we assume the system state \(x \in \mathbb{R}^{n}\) may be separated into so-called \textit{differential} $v \in \mathbb{R}^d$ and \textit{algebraic} $w \in \mathbb{R}^a$ components, such that \(x = (v, w) \in \mathbb{R}^{n}\).
An index-1 DAE may be transformed into an equivalent system of ODEs by differentiating both sides of the algebraic equations $0 = h(v,w,t)$, and rearranging.
We note that the PHDAE (\ref{eq:PHDAE}) can be rewritten in the form of (\ref{eq:DAE}), with differential states $v = \begin{bmatrix} q_C^T & \phi_L^T \end{bmatrix}^T$ and algebraic states $w = \begin{bmatrix} e^T & j_V^T \end{bmatrix}^T$. 

\paragraph{Neural Ordinary Differential Equations}
N-ODEs are a class of deep learning model that uses neural networks to parameterize the right-hand side of an ODE.
That is, neural networks are used to parameterize the unknown components of the time derivative \(f_{\theta}(\cdot)\) of the state \(y\), as in \eqref{eq:NODE}. 
\(f_{\theta}\) is then numerically integrated to obtain predictions of the state at some future time \(k + T\), as in \eqref{eq:NODEint}.
\begin{align}
    \dot{y} &= f_{\theta} (y, t) \label{eq:NODE} \\ 
    y(k+T) &= y(k) + \int_{k}^{k+T} f_{\theta}(y,t) dt \label{eq:NODEint}
\end{align}

\section{Neural Port-Hamiltonian \\ Differential Algebraic Equations} \label{sec:4}
We now present Neural Port-Hamiltonian Differential Algebraic Equations (N-PHDAEs): 
a physics-informed and compositional deep learning approach to modeling dynamical systems subject to algebraic constraints. 
{\color{changes}
As described in \S \ref{sec:intro}, N-PHDAEs may be used to learn control-oriented models of dynamical systems that can be formulated as index-1 DAEs, however, in this work, we focus on their application to modeling electrical networks.
}
\begin{figure*}[t]
    \centering
    \vspace{2mm}
    \includegraphics[]{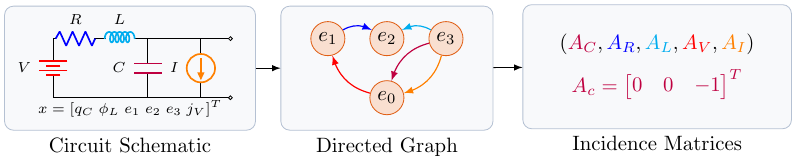}
    \caption{
    Constructing the matrix terms in the N-PHDAE from the interconnection topology of the system components.
    In the context of electrical circuits, a circuit schematic is used to construct a directed graph whose nodes represent junctions and whose edges represent electrical components.
    Component-specific incidence matrices are then obtained from the directed graph, which are in turn used to construct the matrices \(E, J,\) and \(B\) in equation \eqref{eq:PHDescriptor}.
    } 
    \label{fig:incidence_mat}
\end{figure*}

\subsection{Constructing Neural Port-Hamiltonian Differential Algebraic Equations} \label{sec:4.1}

To begin, we assume that the interconnection topology of the system's components is known a priori.
However, models of the individual components are unknown and must be learned from data.
Mathematically, we use the interconnection topology to derive the matrices \(E, J,\) and \(B\) in equation \eqref{eq:PHDescriptor}, and we parameterize the unknown effort \(z(\cdot)\) and dissipation \(r(\cdot)\) functions using neural networks, as illustrated in Figure \ref{fig:PHNDAE}.

In the context of electrical networks, matrices $E$, $J$, and $B$ are expressed in terms of the component-specific incidence matrices $A_C,A_R,A_L,A_V,$ and $A_I$.
Figure \ref{fig:incidence_mat} illustrates the process of extracting these incidence matrices from a circuit schematic.
The procedure begins by transforming the schematic into a directed graph $(\mathcal{V}, \mathcal{E})$.
The nodes $\mathcal{V}$ correspond to the circuit nodes, and the directed edges $\mathcal{E}$ represent the electrical components connecting the nodes (e.g. resistors or capacitors), where the edge direction aligns with the chosen convention for positive current flow. 
The component incidence matrices (\(A_{i}\) for \(i \in \{C, R, L, V, I\}\)) are then obtained by constructing the incidence matrix of the subgraph $(\mathcal{V}, \mathcal{E}_i)$ containing only the edges $\mathcal{E}_i$ corresponding to each component of type \(i\) \citep{gunther2020dynamic}. 
The matrices \(E, J,\) and \(B\) of the N-PHDAE are then defined in terms of the component incidence matrices using equation \eqref{eq:PHDAE}.

We parameterize the N-PHDAE's effort and dissipation functions using the additional physics information that is relevant to these terms in the context of modeling electrical circuits, described by equation \eqref{eq:PHDAE}.
More specifically, \(z_{\theta}(x) := [e, \; \nabla H_{\theta}(\phi_{L}), \; q_{\theta}(q_{C}), \; j_{V}]^{T}\) and \(r_{\theta}(x) := [A_{R}g_{\theta}(A_{R}^{T}e), \; 0, \; A_{C}^{T}e - q_{\theta}(q_{C}), \; 0]^{T}\) where \( g_{\theta}(\cdot) \), \( q_{\theta}(\cdot) \), and \( H_{\theta}(\cdot) \) are parameterized as neural networks.

\subsection{Transforming Neural Port-Hamiltonian DAEs into Systems of Neural ODEs} \label{sec:4.2}
Evaluating and training the constructed N-PHDAEs directly is difficult due to the challenge of solving DAEs in general.
However, under appropriate conditions on the system's interconnection topology \citep{gunther2020dynamic}, the N-PHDAEs that result from \S \ref{sec:4.1} will always be index-1 equations that may be converted into semi-explicit form (as described in \S \ref{sec:background}).

In \S \ref{sec:4.1}, we construct N-PHDAEs in the form of \eqref{eq:PHDescriptor}.
We proceed by automatically identifying the differential \(v \in \mathbb{R}^{d}\) and the algebraic \(w \in \mathbb{R}^{a}\) components of the state \(x\), and by converting the N-PHDAE into semi-explicit form \(\dot{v} = f_{\theta}(v,w, u,t)\) and \(0 = h_{\theta}(v,w, u,t)\),
where \(f_{\theta}(v,w, u,t)\) and \(h_{\theta}(v,w, u,t)\) are functions of \(E, J, B, z_{\theta}(\cdot), r_{\theta}(\cdot)\), and \(u(t)\).
We include derivations for these terms in Appendix \ref{appendix:a}.

After converting the N-PHDAE to semi-explicit form, we use automatic differentiation to transform it into an equivalent system of N-ODEs via index reduction \citep{wanner1996solving}.
\begin{align}
    \begin{bmatrix}
        \dot{v} \\ \dot{w} 
    \end{bmatrix}
    =
    \begin{bmatrix}
        f_{\theta}(\cdot) \\ - \Big(\tfrac{\partial h_{\theta}}{\partial w}\Big)^{-1} \Big(\tfrac{\partial h_{\theta}}{\partial v} f_{\theta}(\cdot) + \tfrac{\partial h_{\theta}}{\partial u} \dot u + \partial_t h_{\theta}(\cdot)\Big)
    \end{bmatrix}
    \label{eq:PHODE}
\end{align}
In \eqref{eq:PHODE}, $\tfrac{\partial h_{\theta}}{\partial v}$, $\tfrac{\partial h_{\theta}}{\partial w}$, and $\tfrac{\partial h_{\theta}}{\partial u}$ denote the Jacobian matrices of $h_{\theta}(\cdot)$ with respect to its input variables, and \(\partial_t h_{\theta}(\cdot)\) is the vector of partial derivatives of \(h_{\theta}(\cdot)\)'s outputs with respect to time \(t\). 
We compute these Jacobian matrices via automatic differentiation, which avoids finite-difference truncation error.
{
\color{changes} We also note that the variables \(v\) and \(w\) in \eqref{eq:PHODE} maintain their original physical interpretations under this transformation.
} 

We include \(\tfrac{\partial h_{\theta}}{\partial u} \dot{u}\) in \eqref{eq:PHODE} for completeness. However, when training and evaluating the N-PHDAE we adopt a zero-order hold on the input: over each integration window \([t, t+T)\), $u(\tau)=u(t)$ for all $\tau$, hence \(\dot u = 0\).

\subsection{Evaluating and Training Neural Port-Hamiltonian Differential Algebraic Equations} \label{sec:4.3}
The N-PHDAE inputs are the state $x(t)$ and control input $u(t)$ at time $t$, and the prediction horizon $T$. 
The model output is the predicted state at time $t+T$, i.e. $\hat{x}(t+T) = \textrm{N-PHDAE}(x, u, t, T)$, obtained by numerically integrating the right-hand side of (\ref{eq:PHODE}) using any ODE solver.

Given a training dataset $\mathcal{D}$, consisting of trajectories $\tau$ of states and control inputs, we optimize the parameters $\theta$ of the N-PHDAE by minimizing the objective \eqref{eq:loss} using gradient-based methods. 
{\color{changes}
}
\begin{align}
    \begin{split}
    \mathcal{L}(\theta, \mathcal{D}) = & \\ \frac{1}{|\mathcal{D}|} \sum_{\tau \in \mathcal{D}} & \sum_{(x, u, t, T, y) \in \tau} \Big[ {\color{black} \alpha} \underbrace{||y - \textrm{N-PHDAE}(x,u,t,T)||_2^2}_{\text{State MSE}} \\
    + &\beta \underbrace{||h_{\theta}(x, u, t)||_2^2}_{\text{Algebraic Eqn. Penalty}}\Big]
     \label{eq:loss}
     \end{split}
\end{align}
Here, the target \(y\) is defined by the true state at the prediction time \(y = x(t+T)\).
The first term of $\mathcal{L}$ ensures the model predictions fit the trajectory data, and the second term of $\mathcal{L}$ encourages \(h_{\theta}(x,u,t)\) to be as close to zero as possible.

{\color{changes}
To evaluate \eqref{eq:PHODE} and train the N-PHDAE, the Jacobian \(\tfrac{\partial h_{\theta}}{\partial w}\) must be invertible.
While this Jacobian is invertible by definition for all index-1 DAEs (see assumption used in \S \ref{sec:4.2}), this is not necessarily the case when \(h_{\theta}(\cdot)\) is parameterized as a neural network that does not perfectly match the true algebraic equations.
However, we found that incorporating the algebraic equation penalty into the loss function was empirically effective in ensuring that \(\tfrac{\partial h_{\theta}}{\partial w}\) remained invertible, thereby maintaining training stability.
}
Furthermore, we observe that training stability is improved when the training procedure begins with a loss that only includes the algebraic equation penalty (i.e., \(\alpha = 0\)), and then switches to include the state MSE loss term after a fixed number of iterations.
This is particularly true when the system includes noisy observations.
A more detailed investigation of methods that strictly enforce the invertibility of \(\tfrac{\partial h_{\theta}}{\partial w}\) is left to future work.
\section{Composing Neural Port-Hamiltonian Differential Algebraic Equations} \label{sec:5}
We now present a method to compose previously learned subsystem N-PHDAEs to obtain an accurate dynamics model for larger composite systems, without additional training.

We define an \textit{interconnection matrix} $A_{\lambda}$ to specify the couplings between an arbitrary number $N$ of pre-defined subsystem N-PHDAEs.
Intuitively, the entries of \(A_{\lambda}\) define couplings between the inputs and outputs of the various subsystems.
In the context of electrical networks, the coupling relations are defined by introducing $n_{\lambda}$ new edges between the nodes of distinct subsystems.
Each new edge models a physical connection in the composite circuit, which may be modeled as a voltage source with a voltage drop of zero and a coupling current \(\lambda\) \citep{gunther2020dynamic}.
The coupling leads to an additional term in Kirchhoff's current law, modeled with an incidence matrix $A_\lambda \in \{-1,0,1\}^{n_{v_c} \times n_{\lambda}}$ describing all the new edges of the composite system, where $n_{v_c} = \sum_{i=1}^{N} n_{v_i}$ is the number of non-grounded nodes in all subsystems.
Given the subsystem N-PHDAEs and the interconnection matrix $A_{\lambda}$, the composite N-PHDAE is defined by
\begin{align}
    \begin{split}
        & \Ec \xc = \\ & \Jc \zc \\ & - \resc + \Bc \controlc,
    \end{split} 
    \label{eq:comp_PHNDAE}
\end{align}
where all subsystem state vectors are concatenated and matrices are stacked diagonally to obtain the corresponding quantities for the composite system. 
\begin{align*}
   s &= \begin{bmatrix} s_1^T & \dots & s_N^T \end{bmatrix}^T, \\ & s \in \{q_C, \phi_L, e, j_V, g_{\theta}(A_R^T e), q_{\theta}(q_C), \nabla H_{\theta}(\phi_L)\}. \\
   A_P &= \text{diag}(A_{P_1}, \dots, A_{P_N}), \; P \in (C,R,L,V,I).
\end{align*}

\begin{figure}[b!]
    \centering
    \includegraphics[width=85.00mm]{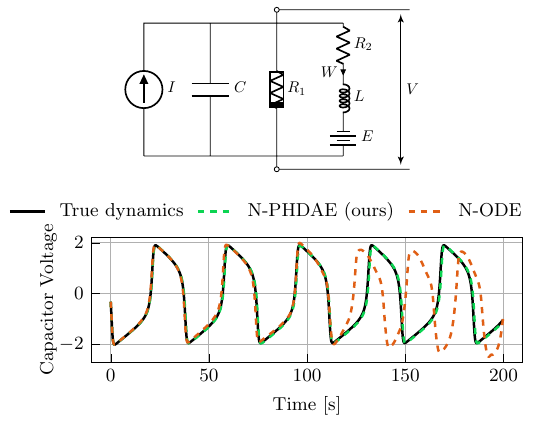}
    \vspace{-5mm}
    \caption{
        Top: The FitzHugh-Nagumo circuit.
        Bottom: Predicted voltage dynamics.
        The baseline N-ODE becomes increasingly inaccurate over long time horizons, while the N-PHDAE maintains accurate predictions.}
    \label{fig:fhn_traj}
\end{figure}

The N-PHDAE of the composite system \eqref{eq:comp_PHNDAE} can then be transformed into equivalent N-ODEs (\S \ref{sec:4.2}). 
The inputs and outputs of the composite N-PHDAE are defined by the concatenation of the subsystem differential states, algebraic states, and control inputs.
\begin{figure}[b!]
    \centering
    \includegraphics[width=85mm]{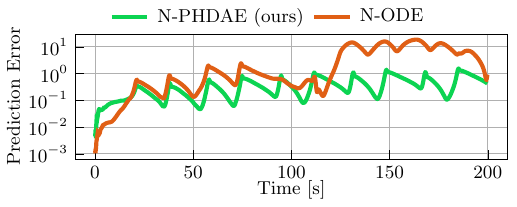}\\
    \includegraphics[width=85mm]{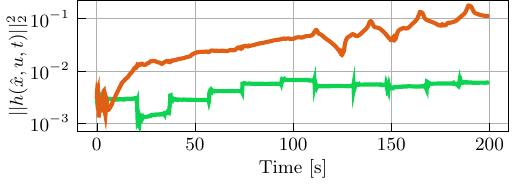}
    \vspace{-4mm}
    \caption{
        Top: The mean square error of the state predictions, as a function of prediction time.
        Bottom: Model violations of true algebraic equations.
        The N-PHDAE model satisfies the ground truth algebraic equations an order of magnitude more effectively than the N-ODE baseline.
    }
    \label{fig:fhn_g_vals}
\end{figure}

\section{Experimental Results} \label{sec:6}

To demonstrate the modeling capabilities of the proposed N-PHDAE, we present two simulation-based case studies.
We begin by training a N-PHDAE model of a well-studied circuit with nonlinear dynamics, before demonstrating the proposed compositional modeling approach via experiments involving interconnected \textit{DC microgrids}.
In all experiments, we parameterize the unknown electrical component relations $g_{\theta}$, $q_{\theta}$ and Hamiltonian $H_{\theta}$ of the N-PHDAE (\S\ref{sec:4.1}) as multi-layer perceptrons. 
We refer the reader to Appendix \ref{appendix:b} {\color{changes}for additional details on experimental setup and hyperparameter selection.}
Project code is available at \href{https://github.com/nathan-t4/NPHDAE}{https://github.com/nathan-t4/NPHDAE}. 

\begin{figure*}[t!]
    \centering    
    \includegraphics{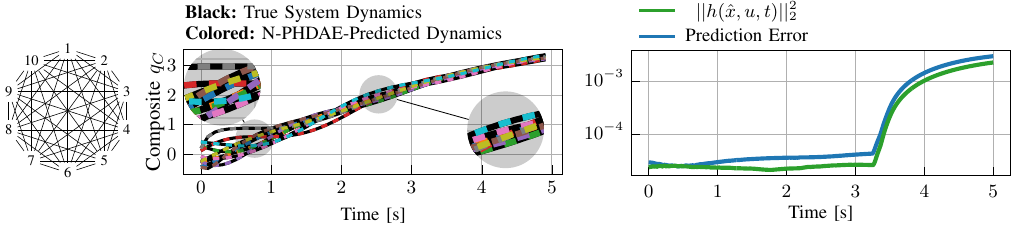}
    \vspace{-2mm}
    \caption{
        Left: The fully connected DC microgrid topology, consisting of 10 interconnected N-PHDAE submodels.
        Middle: The charge of each DGU in the microgrid, as predicted by the composite N-PHDAE.
        Right: The state and constraint error trajectories of the composite DC microgrid.
        The composite model accurately predicts the grid dynamics without additional training.
    }
    \label{fig:dc_microgrid_comp}
\end{figure*}

\begin{figure}[t]
    \centering
    \includegraphics{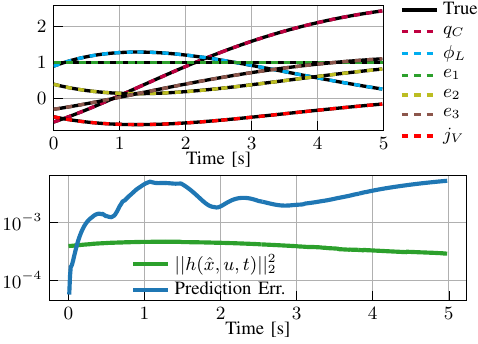}
    \caption{
        Top: N-PHDAE predictions of an individual DGU state trajectory.
        Bottom: Mean squared error, and violation of ground truth algebraic equations, of the predicted trajectory.
    }
    \label{fig:dgu}
\end{figure}

\subsection{Learning the Nonlinear Dynamics of the FitzHugh-Nagumo Circuit}
As an illustrative example, we begin by training a N-PHDAE model of the FitzHugh-Nagumo circuit \citep{Izhikevich:2006}, illustrated in Figure \ref{fig:fhn_traj}, which includes a nonlinear resistor $R_1$.
The training datasets $\mathcal{D}$ consists of 30 state trajectories with randomly sampled initial states (with initial values of the algebraic variables set to ensure feasibility).
To test the robustness of the N-PHDAE training to noise, we also add zero-mean Gaussian random noise with a variance of \(\sigma^{2} = 0.01\) to each observation in the dataset.
We compare the performance of the N-PHDAE to a black-box N-ODE, i.e., a neural ODE that does not leverage physics-based priors, as presented in \S \ref{sec:background}.
The baseline N-ODE is parameterized using an MLP and trained on the same dataset to minimize the mean-squared error of the model's predictions.

\paragraph{N-PHDAEs efficiently learn accurate models of complex dynamics} 
Figure \ref{fig:fhn_traj} illustrates the predictions of the proposed N-PHDAE model over a 200-second horizon.
For comparison, we also illustrate the predictions of a baseline N-ODE. 
The predictions from the N-PHDAE are more accurate than the N-ODE, especially for long horizons. 
This improvement is likely due to the N-PHDAE's ability to better satisfy the system's algebraic constraints, discussed below.

\paragraph{N-PHDAEs learn to satisfy constraint functions orders of magnitude more accurately than the baseline approach} 
Figure \ref{fig:fhn_g_vals} illustrates the N-PHDAE's effectiveness at approximately satisfying the ground truth algebraic equations. 
In particular, we observe that the norm \(||h(\hat{x},u,t)||_{2}^{2}\) of the true algebraic equations evaluated on the model-predicted states is roughly an order of magnitude lower for the N-PHDAE than the N-ODE baseline, resulting in more accurate long-term predictions.

\subsection{Learning Compositional Models of DC Microgrids} \label{sec:6.2}
Next, we demonstrate the compositional approach to learning introduced in \S \ref{sec:5} via experiments simulating a DC microgrid---a low-voltage power grid operating independently from a main grid \citep{cucuzzella2018robust}.
The DC microgrids we consider are powered by distributed generation units (DGUs), such as renewable energy sources, and are interconnected by transmission lines to disperse and store energy. 
We first train a N-PHDAE to model the dynamics of an individual DGU.
The training dataset \(\mathcal{D}\) once again consists of \(30\) trajectories with randomly sampled initial states.
Figure \ref{fig:dgu}, illustrates the N-PHDAE's state predictions and prediction error, as well as a measure of the extent to which the predictions violate constraints. 
As with the FitzHugh-Nagumo circuit, the N-PHDAEs modeling individual DGUs output accurate and constraint-satisfying state predictions.

\paragraph{Compositions of pre-trained N-PHDAE submodels accurately simulate a larger DC microgrid without additional training} 
As illustrated in Figure \ref{fig:Comp} we then compose the learned DGU models by connecting them with transmission lines represented by known PHDAE models.
To do so, we use the composition procedure detailed in \S \ref{sec:5}.
We configure the DC microgrid as a complete graph with 10 nodes, with DGUs as the nodes and transmission lines as the edges. 
In Figure \ref{fig:dc_microgrid_comp}, we plot the trajectories of the capacitor charges $q_C$ within the $10$ interconnected DGUs, as predicted by the composite N-PHDAE representing the entire microgrid. 
We observe that the composite N-PHDAE model outputs accurate predictions for the entire DC microgrid, despite requiring no additional training data from the true composite system itself.

\paragraph{The composite N-PHDAE model continues to satisfy system constraints} 
Despite the interconnected dynamics of the DC microgrid, Figure \ref{fig:dc_microgrid_comp} illustrates that the composition of learned N-PHDAE submodels continues to exhibit small prediction errors and only small violations of the true algebraic constraints.
Indeed, we observe that the magnitude of these values are comparable to those of N-PHDAE submodels representing the individual DGUs, illustrated in Figure \ref{fig:dgu}. 
The proposed N-PHDAEs enable effective compositional approaches to learning models of coupled systems.

\section{Conclusions}
We introduce Neural Port-Hamiltonian Differential Algebraic Equations (N-PHDAEs), a class of physics-informed neural networks that enables compositional approaches to learning system dynamics.
The proposed N-PHDAEs use neural networks to parameterize unknown terms in port-Hamiltonian dynamical systems, even when unknown terms are included in algebraic constraints between state variables.
We propose algorithms for model inference and training that use automatic differentiation to transform the parameterized differential algebraic equations into equivalent systems of Neural Ordinary Differential Equations (N-ODEs), which may be evaluated more easily.
Our experimental results demonstrate that N-PHDAEs learn data-efficient, accurate, and compositional models of electrical networks, enjoying an order of magnitude improvement to constraint satisfaction compared to a baseline N-ODE. 
Future work will examine methods to learn the algebraic coupling constraints between subsystems and apply the proposed compositional learning algorithms to a broader range of constrained and coupled dynamical systems.

\printbibliography


\appendix
\subsection{Transforming the N-PHDAE into Semi-Explicit Form} 
\label{appendix:a}
We explain how to rewrite the N-PHDAE as an index-1 semi-explicit DAE, as described in \S\ref{sec:4.3}. 
First, let the algebraic indices of $E$ be the indices of the rows of $E$ that are all zero, and the differential indices of $E$ as the indices of the rows of $E$ with nonzero elements.
Since $E$ is generally not invertible, we solve a least squares problem to rewrite the N-PHDAE in semi-explicit form. For our experiments, we solve the least-square problem with reduced QR-decomposition \citep{trefethen1997}. 

Let $\bar{E}$ be the submatrix of $E$ that contains the rows corresponding to differential equations (i.e. the rows with non-zero elements) and columns that multiply the differential states. Additionally, let $Q$ and $R$ denote the reduced QR-decomposition of $\bar{E}$, and $\bar{R}$ is the submatrix of $R$ that contains the first $d$ rows of $R$ (where \(d\) is the number of differential state variables). 
The differential equations $f$ of the PHDAE in semi-explicit form may then be expressed as 
\begin{align}
    \dot{v} = \begin{bmatrix} \dot{q}_C \\ \dot{\phi}_L \end{bmatrix} &= f_{\theta}(v,w,u,t) \\
    &= \bar{R}^{\dagger} Q^T (J z_{\theta}(v,w,t) - r_{\theta}(v,w,t) + B u(t)).
\end{align}
The algebraic equations $h_{\theta}(\cdot)$ correspond to the remaining rows of the right-hand side of the N-PHDAE, i.e., the entries of the vector output of \(J z_{\theta}(v,w,t) - r_{\theta}(v,w,t) + B u(t)\) that share indices with the algebraic variables.
\begin{align}
    0 &= h_{\theta}(v,w,u,t) \\
    &= [J z_{\theta}(v,w,t) - r_{\theta}(v,w,t) + B u(t)]_{\textrm{algIndices}}.
\end{align}
\subsection{Additional Experimental Details}
\label{appendix:b}
The code to reproduce all numerical experiments is implemented in Python using the Jax \citep{jax2018github} and Haiku \citep{haiku2020github} libraries and available at \href{https://github.com/nathan-t4/NPHDAE}{https://github.com/nathan-t4/NPHDAE}. 
All numerical experiments are trained for $100000$ epochs with a batch size of $128$, loss function hyper-parameter $\alpha=0.01$ (\S\ref{sec:4.3}), and optimized using Adam. The learning rate is initially set to $0.0001$ for the N-PHDAE and $0.001$ for the baseline black-box NODE, and decays to zero with cosine annealing \citep{loshchilov2016sgdr}. 
These learning rates were chosen using a coarse hyperparameter sweep to maximize performance of each of the tested algorithms, respectively.
The remaining training hyperparameters are fixed between the different approaches to ensure a fair comparison between algorithms.
The hyperparameter values reported above were selected to yield strong performance across both approaches.
The resistor and capacitor component relations $g_{\theta}$, $q_{\theta}$ and Hamiltonian $H_{\theta}$ of the N-PHDAE and the baseline black-box NODE are all parameterized with multi-layer perceptions, each with two hidden layers of 100 nodes and ReLU activation.

The training datasets are generated by rewriting the electrical network dynamics as a Port-Hamiltonian differential algebraic equation (\ref{eq:PHDAE}). Then, the Port-Hamiltonian differential algebraic equation is transformed to an equivalent ordinary differential equation using index reduction (\S\ref{sec:4.2}), and the state at the next time-step is obtained through numerical integration using the fourth-order Runge-Kutta method with a fixed time-step (\S\ref{sec:4.3}). For all numerical experiments, we generate $30$ trajectories of $1000$ time-steps for the training dataset and $10$ trajectories of $10000$ time-steps for the validation dataset {\color{changes} with constant \(\Delta t = 0.01\)}. To test the robustness of the N-PHDAE to noise, we add zero-mean Gaussian random noise with variance $\sigma^2=0.01$ to the training data for the FitzHugh-Nagumo circuit simulation. 

Empirically, we found that switching the loss function (Equation \ref{eq:loss}) after a fixed number of epochs improves the training performance of N-PHDAEs. We set $\alpha = 0$ and $\beta = 1$ for the first 25000 epochs to encourage the model to minimize the constraint violation. Then, for the remaining 75000 epochs, we set $\alpha = 1$ and $\beta = 0.01$ to minimize the mean-squared error over all the states. 

\subsection{FitzHugh-Nagumo Circuit}
The FitzHugh-Nagumo model is a well-studied nonlinear dynamics model of excitable biological systems first introduced by \cite{fitzhugh1961}, and with an equivalent circuit derived by \cite{nagumo1962}. The governing equations of the membrane potential $V$ and recovery variable $W$ are provided below, where $I$ is the stimulus current. 
\begin{align}
    \begin{split}
        \dot{V} &= V - V^3/3 - W + I \\
        \dot{W} &= 0.08(V + 0.7 - 0.8W)
    \end{split}
    \label{eq:FHN}
\end{align}
The equivalent circuit representation shown in Figure (\ref{fig:fhn_traj}) has parameter values of $R_2=0.8$, $L=1/0.08$, $C=1.0$, $E=-0.7$, and $I=1.0$. The capacitor voltage represents the membrane potential $V$, and the inductor current represents the recovery variable $W$. The electrical component values are set to match the governing equations of the circuit to (\ref{eq:FHN}). The equivalent PHDAE of the FitzHugh-Nagumo circuit (\ref{eq:PHDAE}) has incidence matrices:
\begin{gather}
    \begin{split}
    A_C = \begin{bmatrix}
        1 \\ 0 \\ 0
    \end{bmatrix}, \
    A_R = \begin{bmatrix}
        1 & -1 \\
        0 & 1 \\
        0 & 0
    \end{bmatrix}, \
    A_L = \begin{bmatrix}
        0 \\ 0 \\ -1
    \end{bmatrix}, \\
    A_V = \begin{bmatrix}
        0 \\ -1 \\ 1
    \end{bmatrix}, \
    A_I = \begin{bmatrix}
        1 \\ 0 \\ 0
    \end{bmatrix}
    \end{split}
\end{gather}

and known component relations:
\begin{align}
    \begin{split}
        r&\colon \begin{bmatrix} V_{R1} & V_{R2} \end{bmatrix} \mapsto \begin{bmatrix} V_{R1}^3 \ / \ 3 - V_{R1} & V_{R2} \ / \ R_2\end{bmatrix} \\
        q&\colon V_{C} \mapsto C V_{C} \\
        H&\colon \phi_L \mapsto \phi_L^2 \ / \ 2L
    \end{split}
\end{align}
The initial conditions for $V$ and $W$ for the training datasets are sampled from the uniform distribution $U(-3.0, 3.0)$ with $\Delta t = 0.1$. The baseline black-box neural differential ordinary equation (NODE) is trained on the same dataset as the N-PHDAE, and is trained to optimize the mean-squared error on the state.
\begin{align}
    \mathcal{L}(\omega, t) = \frac{1}{|\mathcal{D}|} \sum_{\tau \in \mathcal{D}} \sum_{(\omega, u, t) \in \tau} ||\omega - \NODE(\omega)||_2^2
\end{align}
Here $\omega$ is the state and $u$ is the control input.

\subsection{Microgrids}
We use the direct current (DC) microgrid model introduced by \cite{cucuzzella2018robust} for the compositional learning experiment.
DC microgrids are small-scale power grids composed of distributed generation units, loads, and energy storage systems. 

\subsubsection{Distributed Generation Unit Model}
Distributed generation units (DGU) are small-scale electricity generators that are an alternative to traditional power plants. For example, renewable energy sources can act as the power generation unit in the DGU model. The equivalent PHDAE of the distributed generation unit (\ref{eq:PHDAE}) has incidence matrices:
\begin{gather}
    \begin{split}
    A_C = \begin{bmatrix}
        0 \\ 0 \\ 1
    \end{bmatrix}, \
    A_R = \begin{bmatrix}
        -1 \\ 1 \\ 0
    \end{bmatrix}, \
    A_L = \begin{bmatrix}
        0 \\ 1 \\ -1
    \end{bmatrix}, \\
    A_V = \begin{bmatrix}
        1 \\ 0 \\ 0
    \end{bmatrix}, \
    A_I = \begin{bmatrix}
        0 \\ 0 \\ -1
    \end{bmatrix}
    \end{split}
\end{gather}
and component relations:
\begin{align}
    \begin{split}
        r&\colon V_R \mapsto V_R \ / \ R \\
        q&\colon V_{C} \mapsto C V_{C} \\
        H&\colon \phi_L \mapsto \phi_L^2 \ / \ 2L
    \end{split}
\end{align}
The training dataset for the distributed generation unit has $\Delta t = 0.01$ and parameter values $R_{dgu}=1.2$, $L_{dgu}=1.8$, and $C_{dgu}=2.2$. 

\subsubsection{Transmission Line Model}
The transmission lines interconnect two distributed generation units and model the grid loads. The equivalent PHDAE of the transmission line model has incidence matrices:
\begin{gather}
\begin{split}
    A_C = \begin{bmatrix}
        0 \\ 0 \\ 0
    \end{bmatrix}, \
    A_R = \begin{bmatrix}
        1 \\ -1 \\ 0
    \end{bmatrix}, \
    A_L = \begin{bmatrix}
        0 \\ -1 \\ 1
    \end{bmatrix}, \\
    A_V = \begin{bmatrix}
        0 \\ 0 \\ 0
    \end{bmatrix}, \
    A_I = \begin{bmatrix}
        0 \\ 0 \\ 0
    \end{bmatrix}
\end{split}
\end{gather}
and component relations:
\begin{align}
    \begin{split}
        r&\colon V_R \mapsto V_R \ / \ R_{tl} \\
        H&\colon \phi_L \mapsto \phi_L^2 \ / \ 2L_{tl}
    \end{split}
\end{align}
The transmission line models have parameter values $R_{tl}$ and $L_{tl}$ sampled from the uniform distribution $U(0.1, 2.0)$.

\subsubsection{Obtaining the Microgrid with Composition}
Towards simulating microgrids via composition, we need an interconnection matrix $A_{\lambda}$ which specifies how the various distributed generation units and transmission lines are interconnected. In the compositional learning experiments, the microgrid has a complete graph configuration with 10 nodes, with DGUs at the nodes and transmission lines at the edges. 
Due to space constraints, we do not include $A_\lambda \in \{-1,0,1\}^{165 \times 90}$ for the microgrid configuration as a complete graph with 10 nodes. Instead, for illustration purposes, we include $A_{\lambda}$ when the microgrid is configured as a complete graph with 2 nodes:
\begin{align}
    A_{\lambda} = \begin{bmatrix}
        0 & 0 \\ 0 & 0 \\ 1 & 0 \\ 0 & 0 \\ 0 & 0 \\ 0 & -1 \\ -1 & 0 \\ 0 & 0 \\ 0 & 1
    \end{bmatrix}
\end{align}
We can then derive the composite N-PHDAE by stacking the vectors, i.e. $q = \begin{bmatrix} q_1^T & q_2^T \end{bmatrix}$ and $A_C = \text{diag}(A_{C1}, A_{C2})$. However, we emphasize that our choice of the microgrid configuration is arbitrary; we can simulate a microgrid with any configuration using the compositional learning framework introduced in \S\ref{sec:5} if given the corresponding interconnection matrix $A_{\lambda}$. 

\end{document}